\documentclass[aoas,preprint]{imsart}

\usepackage{fullpage}
\setattribute{journal}{name}{}

\usepackage{amsmath}
\usepackage{amssymb}
\usepackage{graphicx}
\usepackage{algpseudocode}
\usepackage{algorithm}
\usepackage{todonotes}
\usepackage{subfig}
\usepackage{booktabs}

\usepackage{natbib}
\setcitestyle{round}

\renewcommand{\tt}[1]{\texttt{#1}} 

\newcommand{\ra}{\rightarrow}
\newcommand{\qq}{q_{d\ra b}}
\renewcommand{\th}{\theta}
\newcommand{\given}{\,|\,}
\newcommand{\trans}{\textsf{T}}
\let\originaleqref\eqref
\renewcommand{\eqref}{Equation~\originaleqref}

\begin{document}

\begin{frontmatter}

\title{Firefly Monte Carlo: Exact MCMC with Subsets of Data}
\runtitle{Firefly Monte Carlo}

\begin{aug}
\author{\fnms{Dougal} \snm{Maclaurin}\ead[label=e1]{maclaurin@physics.harvard.edu}}
\and
\author{\fnms{Ryan P.} \snm{Adams}\ead[label=e2]{rpa@seas.harvard.edu}}

\affiliation{Harvard University}

\runauthor{Maclaurin and Adams}

\address{Dougal Maclaurin\\
Department of Physics\\
Harvard University\\
Cambridge, MA\\
\printead{e1}}

\address{Ryan P. Adams\\
School of Engineering and Applied Sciences\\
Harvard University\\
Cambridge, MA\\
\printead{e2}}
\end{aug}

\begin{abstract}
Markov chain Monte Carlo (MCMC) is a popular and successful general-purpose tool for Bayesian inference.  However, MCMC cannot be practically applied to large data sets because of the prohibitive cost of evaluating every likelihood term at every iteration. Here we present \emph{Firefly Monte Carlo} (FlyMC) an auxiliary variable MCMC algorithm that only queries the likelihoods of a potentially small subset of the data at each iteration yet simulates from the exact posterior distribution, in contrast to recent proposals that are approximate even in the asymptotic limit.  FlyMC is compatible with a wide variety of modern MCMC algorithms, and only requires a lower bound on the per-datum likelihood factors.  In experiments, we find that FlyMC generates samples from the posterior more than an order of magnitude faster than regular MCMC, opening up MCMC methods to larger datasets than were previously considered feasible.
\end{abstract}

\end{frontmatter}

\section{INTRODUCTION}

The Bayesian approach to probabilistic modeling is appealing for a several reasons: the generative framework allows one to separate out modeling assumptions from inference procedures, outputs include estimates of uncertainty that can be used for decision making and prediction, and it provides clear ways to perform model selection and complexity control. Unfortunately, the fully-Bayesian approach to modeling is often very computationally challenging.  It is unusual for non-trivial models of real data to have closed-form posterior distributions.  Instead, one uses approximate inference via Monte Carlo, variational approximations, Laplace approximations, or other tools.

One of the persistent challenges to Bayesian computation is that coherent procedures for inference appear to require examination of all of the data in order to evaluate a new hypothesis regarding parameters or latent variables.  For example, when performing Metropolis--Hastings (MH), it is necessary to evaluate the target posterior density for each proposed parameter update, and this posterior will usually contain a factor for each datum.  Similarly, typical variational Bayesian procedures need to build local approximations for each of the data in order to update the approximation to any global parameters.  In both cases, it may be necessary to perform these data-intensive computations many times as part of an iterative procedure.

Recent methods have been proposed to partially overcome these difficulties.  Stochastic and online variational approximation procedures \citep{hoffman-etal-2010a,Hoffman2013} can use subsets of the data to make approximations to global parameters.  As these procedures are optimizations, it is possible to build on convergence results from the stochastic optimization literature and achieve guarantees on the resulting approximation.  For Markov chain Monte Carlo, the situation is somewhat murkier.  Recent work has considered how subsets of data might be used to approximate the ideal transition operator for Metropolis--Hastings (MH) and related algorithms \citep{Welling2011}. \citet{Korattikara2014} and \citet{Bardenet2014} have recently shown that such approximate MH moves can lead to stationary distributions which are approximate but that have bounded error, albeit under strong conditions of rapid mixing.

In this paper, we present a Markov chain Monte Carlo algorithm, \emph{Firefly Monte Carlo} (FlyMC), that is in line with these latter efforts to exploit subsets of data to construct transition operators.  What distinguishes the approach we present here, however, is that this new MCMC procedure is \emph{exact} in the sense that it leaves the true full-data posterior distribution invariant.  FlyMC is a latent variable model which introduces a collection of Bernoulli variables -- one for each datum -- with conditional distributions chosen so that they effectively turn on and off data points in the posterior, hence the ``firefly'' name.  The introduction of these latent variables does not alter the marginal distribution of the parameters of interest.  Our only requirement is that it be possible to provide a ``collapsible'' lower bound for each likelihood term.  FlyMC can lead to dramatic performance improvements in MCMC, as measured in wallclock time.

The paper is structured as follows.  In Section~\ref{sec:firefly}, we introduce Firefly Monte Carlo and show why it is valid.  Section~\ref{sec:implementation} discusses practical issues related to implementation of FlyMC.  Section~\ref{sec:experiments} evaluates the new method on several different problems, and Section~\ref{sec:discussion} discusses its limitations and possible future directions.

\section{FIREFLY MONTE CARLO}
\label{sec:firefly}

The Firefly Monte Carlo algorithm tackles the problem of sampling from the posterior distribution of a probabilistic model.  We will denote the parameters of interest as~$\th$ and assume that they have prior~$p(\theta)$.  We assume that~$N$ data have been observed~$\{x_n\}^N_{n=1}$ and that these data are conditionally independent given~$\theta$ under a likelihood~$p(x_n\given\theta)$.  Our target distribution is therefore
\begin{align}
p(\theta\,|\{x_n\}^N_{n=1}) &\propto p(\theta, \{x_n\}^N_{n=1})
= p(\theta)\!\prod^N_{n=1}\! p(x_n|\theta).
\label{SimplePosterior}
\end{align}
For notational convenience, we will write the~$n$th likelihood term as a function of~$\theta$ as
\begin{align*}
    L_n(\theta) &= p(x_n\given\theta)\,.
\end{align*}
An MCMC sampler makes transitions from a given $\th$ to a new $\th'$ such that posterior distribution remains invariant. Conventional algorithms, such as Metropolis--Hastings, require evaluation of the unnormalized posterior in full at every iteration. When the data set is large, evaluating all~$N$ likelihoods is a computational bottleneck. This is the problem that we seek to solve with FlyMC.

For each data point, $n$, we introduce a binary auxiliary variable,~${z_n \in \{0,1\}}$, and a function~$B_n(\th)$ which is a sctrictly positive lower bound on the~$n$th likelihood:~${0 < B_n(\th) \leq L_n(\th)}$. Each~$z_n$ has the following Bernoulli distribution conditioned on the parameters:
\begin{align*}
    p(z_n\given x_n, \th) &= \left [\frac{L_n(\th) - B_n(\th)}{L_n(\th)}  \right ]^{z_n}\left [\frac{B_n(\th)}{L_n(\th)}  \right ]^{1-z_n}.
\end{align*}
We now augment the posterior distribution with these~$N$ variables:
\begin{align*}
 p(\theta, \{z_n\}^N_{n=1}\given\{x_n\}^N_{n=1})&\propto p(\theta, \{x_n, z_n\}^N_{n=1}) \\
&= p(\theta)\prod^N_{n=1} p(x_n\given\theta)\,p(z_n\given x_n, \theta)\,.
\end{align*}
As in other auxiliary variable methods such as slice sampling, Swendsen-Wang, or Hamiltonian Monte Carlo, augmenting the joint distribution in this way does not damage the original marginal distribution of interest:
\begin{align*}
    &\quad \sum_{z_1} \cdots \sum_{z_N}p(\theta)\prod^N_{n=1}\,p(x_n\given\theta)\,p(z_n\given x_n,\theta)\\
    &= p(\theta)\,\prod^N_{n=1}p(x_n\given\theta)\sum_{z_n}p(z_n\given x_n,\theta)\\
    &= p(\theta)\,\prod^N_{n=1}p(x_n\given\theta)
\end{align*}
However, this joint distribution has a remarkable property: to evaluate the probability density over~$\theta$, given a particular configuration of~$\{z_n\}^N_{n=1}$, it is only necessary to evaluate those likelihood terms for which~${z_n=1}$.  Consider factor~$n$ from the product above:
\begin{align*}
& \quad p(x_n\given \theta)p(z_n\given x_n,\theta)\\
&= L_n(\theta)\left [\frac{L_n(\th) - B_n(\th)}{L_n(\th)}  \right ]^{z_n}\left [\frac{B_n(\th)}{L_n(\th)}  \right ]^{1-z_n}\\
&= \begin{cases}
L_n(\theta) - B_n(\theta) & \text{if $z_n=1$}\\
B_n(\theta) & \text{if $z_n=0$}
\end{cases}\,.
\end{align*}
The ``true'' likelihood term~$L_n(\theta)$ only appears in those factors for which~${z_n=1}$ and we can think of these data as forming a ``minibatch'' subsample of the full set.  If most~${z_n=0}$, then transition updates for the parameters will be much cheaper, as these are applied to~$p(\theta\given \{x_n,z_n\}^N_{n=1})$.

Of course, we do have to evaluate all~$N$ bounds $B_n(\th)$ at each iteration.  At first glance, we seem to have just shifted the computational burden from evaluating the $L_n(\th)$ to evaluating the $B_n(\th)$.  However, if we choose $B_n(\th)$ to have a convenient form, a scaled Gaussian or other exponential family distribution, for example, then the full product~$\prod_{n=1}^N B_n(\th)$ can be computed for each new $\theta$ in $O(1)$ time using the sufficient statistics of the distribution, which only need to be computed once.  To make this clearer, we can rearrange the joint distribution in terms of a ``pseudo-prior,'' $\tilde p(\th)$ and ``pseudo-likelihood,'' $\tilde L_n(\th)$ as follows:
\begin{align}
p(\th, \{z_n\}^{N}_{n=1} \given \{x_n\}^N_{n=1})
&\propto \tilde{p}(\theta)\prod_{n:z_n=1}\tilde{L}_n(\theta)
\label{JointDist}
\end{align}
where the product only runs over those $n$ for which $z_n=1$, and we have defined 
\begin{align*}
\tilde p(\th) &= p(\th) \prod_{n=1}^N B_n(\th) &
\tilde L_n(\th) &= \frac{L_n(\th)-B_n(\th)}{B_n(\th)}\,.
\end{align*}

We can generate a Markov chain for the joint distribution in \eqref{JointDist} by alternating between updates of $\th$ conditional on~$\{z_n\}^N_{n=1}$, which can be done with any conventional MCMC algorithm, and updates of~$\{z_n\}^N_{n=1}$ conditional on $\th$ for which we discuss efficient methods in Section~\ref{Resampling}. We emphasize that the marginal distribution over $\th$ is still the correct posterior distribution given in \eqref{SimplePosterior}.

\begin{figure}[t]
\includegraphics[clip=true,trim=0 15 0 0]{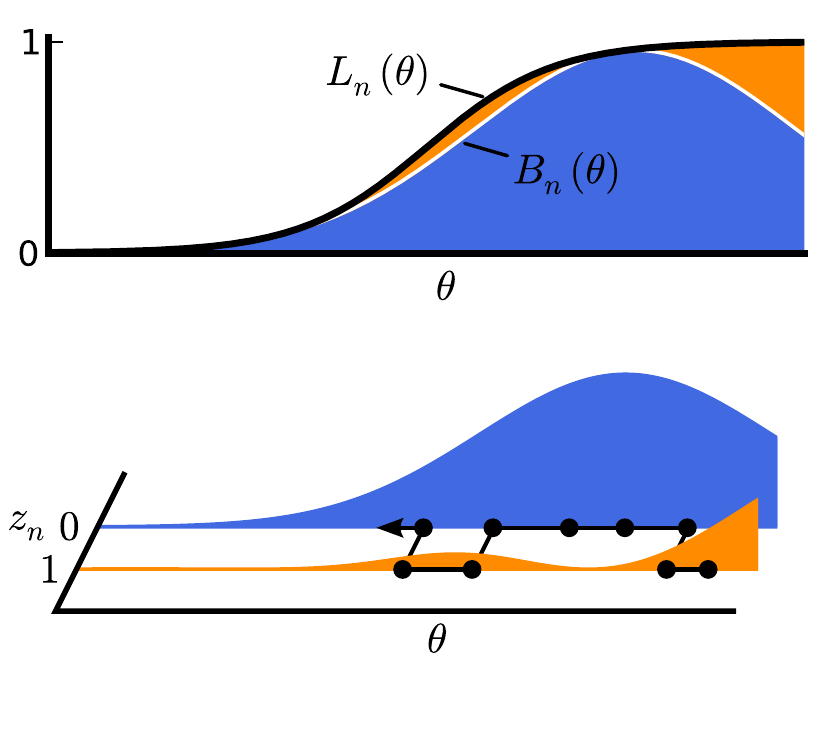}
\caption{Illustration of the auxiliary variable representation of a single likelihood for a one-dimensional logistic regression model.  The top panel shows how the likelihood function, $L_n(\th)$, corresponding to a single datum $n$, can be partitioned into two parts: a lower bound, $B_n(\th)$, shaded blue, and the remainder, shaded orange. The bottom panel shows that we can introduce a Bernoulli random variable $z_n$ and construct a Markov chain in this new, higher dimensional space, such that marginalizing out (i.e. ignoring) the $z_n$ recovers the original likelihood. If $B_n(\th) \gg L_n(\th)-B_n(\th)$, the Markov chain will tend to occupy $z_n=0$ and we can avoid evaluating $L_n(\th)$ at each iteration.}
\label{Cartoon}
\end{figure}

At a given iteration, the ${z_n=0}$ data points are ``dark'': we simulate the Markov chain without computing their likelihoods. Upon a Markov transition in the space of~$\{z_n\}^N_{n=1}$, a smattering of these dark data points become ``bright'' with their ${z_n=1}$, and we include their likelihoods in subsequent iterations. The evolution of the chain evokes an image of fireflies, as the individual data blink on and off due to updates of the~$z_n$.

\begin{figure}[t!]
\includegraphics[clip=true,trim=0 0 0 0]{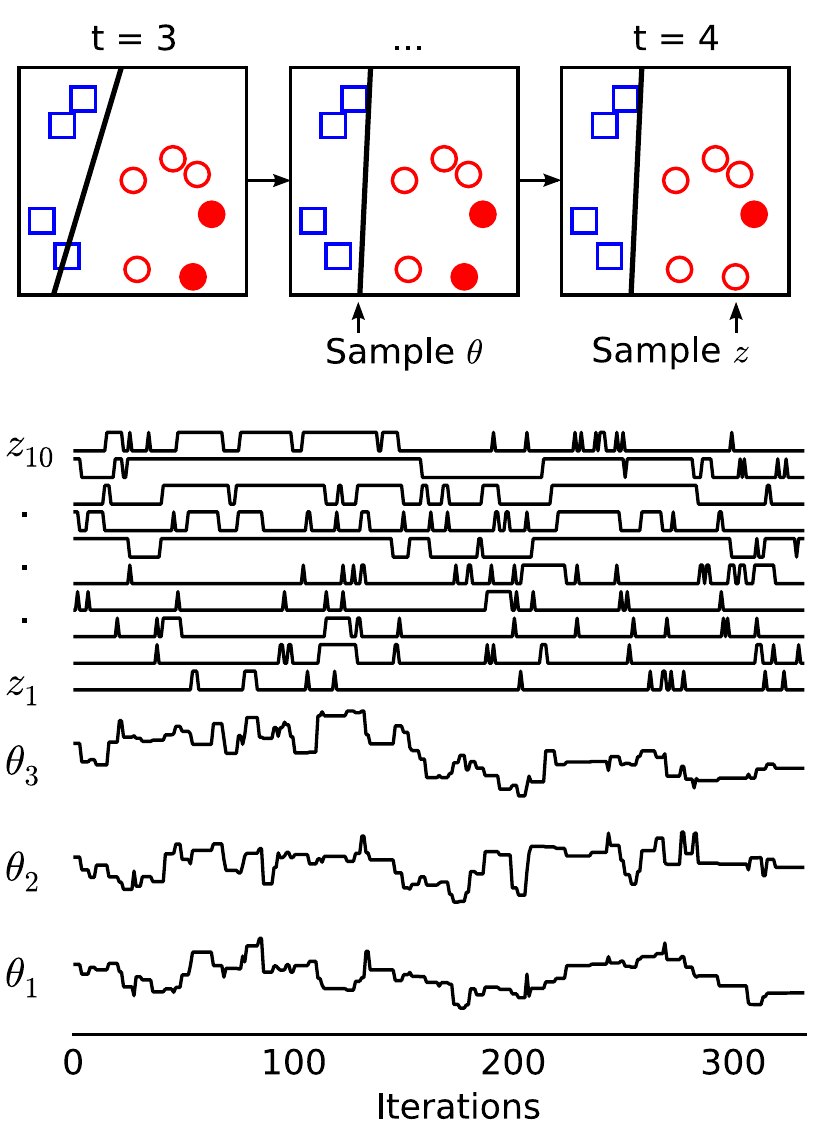}
\caption{Illustration of the FlyMC algorithm operating on a logistic regression model of a toy synthetic data set, a two-class classification problem in two dimensions (and one bias dimension).  The top panel shows a single iteration of FlyMC, from $t=3$ to $t=4$, which consists of two steps: first we sample $\theta$, represented by the line of equal class probability. Next we sample the $z_n$. In this case, we see one `bright' (solid) data point become dark. The bottom panel shows the trajectories of all components of $\theta$ and $z$.}
\label{toy_model}
\end{figure}

The details of choosing a lower bound and efficiently sampling the $\{z_n\}$ are treated in the proceeding sections, but the high-level picture is now complete. Figure \ref{Cartoon} illustrates the augmented space, and a simple version of the algorithm is shown in Algorithm~\ref{basic_flymc}.  Figure~\ref{toy_model} shows several steps of Firefly Monte Carlo on a toy logistic regression model.

\section{IMPLEMENTATION CONSIDERATIONS}
\label{sec:implementation}

In this section we discuss two important practical matters for implementing an effective FlyMC algorithm: how to choose and compute lower bounds, and how to sample the brightness variables $z_n$. For this discussion we will assume that we are dealing with a data set consisting of~$N$ data points, and a parameter set,~$\th$, of dimension~${D\ll N}$. We will also assume that it takes at least~$O(ND)$ time to evaluate the likelihoods at some~$\th$ for the whole data set and that evaluating this set of likelihoods at each iteration is the computational bottleneck for MCMC. We will mostly assume that space is not an issue: we can hold the full data set in memory and we can afford additional data structures occupying a few bytes for each of the~$N$ data.

The goal of an effective implementation of FlyMC is to construct a Markov chain with similar convergence and mixing properties to that of regular MCMC, while only evaluating a subset of the data points on average at each iteration.  If the average number of ``bright'' data points is~$M$, we would like this to achieve a computational speedup of nearly $N/M$ over regular MCMC.

\subsection{Choosing a lower bound}

The lower bounds, $B_n(\th)$ of each data point's likelihood $L_n(\th)$ should satisfy two properties. They should be relatively tight, and it should be possible to efficiently summarize a product of lower bounds $\prod_{n} B_n(\th)$ in a way that (after setup) can be evaluated in time independent of~$N$.

The tightness of the bounds is important because it determines the number of bright data points at each iteration, which determines the time it takes to evaluate the joint posterior.  For a burned-in chain, the average number of bright data points, $M$, will be:
\begin{equation*}
M = \sum_{n=1}^N \langle z_n \rangle = \sum_{n=1}^N \int\!p(\th\given\{x_n\}^N_{n=1}) \frac{L_n(\th)\!-\!B_n(\th)}{L_n(\th)}\mathrm{d}\theta\,.
\end{equation*}
Therefore it is important that the bounds are tight at values of $\th$ where the posterior puts the bulk of its mass.

The second important property is that the product of the lower bounds must be easy to compute and represent. This property emerges naturally if we use scaled exponential-family lower bounds so that their product can be summarized via a set of sufficient statistics.  We should also mention that the individual bounds~$B_n(\th)$ should be easy to compute themselves, since these are computed alongside $L_n(\th)$ for all the bright points at each iteration. In all the examples considered in this paper, the rate-limiting step in computing either $L_n(\th)$ or~$B_n(\th)$ is the evaluation of the dot product of a feature vector with a vector of weights.  Once we have computed~$L_n(\th)$ the extra cost of computing~$B_n(\th)$ is negligible.

\begin{algorithm*}[t!]
\caption{Firefly Monte Carlo \hfill \textit{Note: Using simple random-walk MH for clarity.}}
\label{basic_flymc}
  \begin{algorithmic}[1]
    \State $\theta_0 \sim \textsc{InitialDist}$
        \Comment{Initialize the Markov chain state.}

    \For{$i\gets 1\ldots\textsc{Iters}$}
        \Comment{Iterate the Markov chain.}
        
        \For{$j\gets 1\ldots\lceil N \times \textsc{ResampleFraction} \rceil$}\label{alg:fmc:z_start}
            \State $n \sim \text{RandInteger}(1,N)$
                \Comment{Select a random data point.}
            \State $z_n \sim \text{Bernoulli}(1 - B_n(\theta_{i-1})/L_n(\theta_{i-1}))$
                \Comment{Biased coin-flip to determine whether $n$ is bright or dark.}
        \EndFor\label{alg:fmc:z_end}
    
    \State $\theta' \gets \theta_{i-1} + \eta$ where $\eta \sim \text{Normal}(0,\epsilon^2\mathbb{I}_D)$
        \Comment{Make a random walk proposal with step size $\epsilon$.}
    \State $u \sim \text{Uniform}(0,1)$
        \Comment{Draw the MH threshold.}
    \If{ $\displaystyle\frac{ \textsc{JointPosterior}(\theta'\,;\,\{z_n\}^N_{n=1})}
    {\textsc{JointPosterior}(\theta\,;\,\{z_n\}^N_{n=1})} > u$ }
        \Comment{Evaluate MH ratio conditioned on auxiliary variables.}
        \State $\theta_i \gets \theta'$
            \Comment{Accept proposal.}
    \Else
        \State $\theta_i \gets \theta_{i-1}$
            \Comment{Reject proposal and keep current state.}
    \EndIf
    \EndFor
    \State
    \Function{JointPosterior}{$\th \,;\,\{z_n\}^N_{n=1}$}
        \Comment{Modified posterior that conditions on auxiliary variables.}
      \State $P \gets p(\th)\times\prod^N_{n=1}B_n(\theta)$
          \Comment{Evaluate prior and bounds.  Collapse of bound product not shown.}      
      \For {\textbf{each} $n$ \textbf{for which} ${z_n = 1}$}
          \Comment{Loop over bright data only.}
        \State $P \gets P \times (L_n(\th)/B_n(\th)-1)$
            \Comment{Include bound-corrected factor.}
      \EndFor
      \State \textbf{return} $P$
    \EndFunction

  \end{algorithmic}
\end{algorithm*}

At this stage it is useful to consider a concrete example. The logistic regression likelihood is
\begin{equation*}
L_n(\th) = \text{logit}^{-1}(t_n\th^{\trans} x_n) = \frac{1}{1 + \exp\{-t_n\th^{\trans} x_n\}}\,,
\end{equation*}
where ${x_n \in \mathbb{R}^D}$ is the set of features for the $n$th data point and ${t_n \in \{-1, 1\}}$ is its class. The logistic function has a family of scaled Gaussian lower bounds, described in \citet{Jaakkola1997}, parameterized by~$\xi$, the location at which the bound is tight:
\begin{align*}
\log(B_n(\th)) &= a (t_n\th^{\trans} x_n)^2 + b(t_n\th^{\trans} x_n) + c
\end{align*}
where:
\begin{eqnarray*}
a &=& \frac{-1}{4\xi}\left(\frac{e^{\xi}-1}{e^{\xi} + 1}\right)
\qquad b = \frac{1}{2} \\
c &=& -a*\xi^2 + \frac{\xi}{2} - \log\left (e^{\xi} + 1\right )
\end{eqnarray*}

This is the bound shown in Fig. \ref{Cartoon}.  The product of these bounds can be computed for a given $\th$ in $O(D^2)$ time, provided we have precomputed the moments of the data, at a one-time setup cost of $O(ND^2)$:
\begin{align*}
\frac{1}{N}\log \prod_{n=1}^N B_n(\th)
= a \th^{\trans}\hat{S}\theta + b\theta^{\trans}\hat{\mu} + c
\end{align*}
where
\begin{align*}
\hat{S} &= \frac{1}{N}\sum_{n=1}^N x_n x_n^{\trans} &
\hat{\mu} &= \frac{1}{N}\sum_{n=1}^N t_n x_n\,.
\end{align*}

This bound can be quite tight. For example, if we choose~${\xi = 1.5}$ the probability of a data point being bright is less than 0.02 in the region where~${0.1 < L_n(\th) < 0.9}$. With a bit of up-front work, we can do even better than this by choosing bounds that are tight in the right places. For example, we can perform a quick optimization to find an approximate maximum \emph{a posteriori} (MAP) value of $\th$ and construct the bounds to be tight there. We explore this idea further in Section~\ref{sec:experiments}.

\subsection{Sampling and handling the auxiliary brightness variables}
\label{Resampling}

The resampling of the~$z_n$ variables, as shown in lines \ref{alg:fmc:z_start} to \ref{alg:fmc:z_end} of Algorithm~\ref{basic_flymc}, takes a step by explicitly sampling~$z_n$ from its conditional distribution for a random fixed-size subset of the data.  We call this approach \emph{explicit resampling} and it has a clear drawback: if the fixed fraction is~$\alpha$ (shown as \textsc{ResampleFraction} in Algorithm~\ref{basic_flymc}), then the chain cannot have a mixing time faster than $1/\alpha$, as each data point is only visited a fraction of the time.

Nevertheless, explicit resampling works well in practice since the bottleneck for mixing is usually the exploration of the space of~$\th$, not space of~$z_n$. Explicit resampling has the benefit of being a simple, low-overhead algorithm that is easy to vectorize for speed. The variant shown in Algorithm \ref{basic_flymc} is the simplest: data points are chosen at random, with replacement. We could also sample without replacement but this is slightly harder to do efficiently.  Another variant would be to deterministically choose a subset from which to Gibbs sample at each iteration. This is more in line with the traditional approach of stochastic gradient descent optimization. Such an approach may be appropriate for data sets which are too large to fit into memory, since we would no longer need random access to all data points.  The resulting Markov chain would be non-reversible, but still satisfy stationarity conditions.

Explicitly sampling a subset of the $z_n$ seems wasteful if~${M \ll N}$, since most updates to~$z_n$ will leave it unchanged. We can do better by drawing each update for~$z_n$ from a pair of tunable Bernoulli proposal distributions~${q(z'_n=1\given z_n=0) = q_{d\to                     b}}$ and~${q(z'_n=0\given z_n=1) = q_{b\to d}}$,  and then performing a Metropolis--Hastings accept/reject step with the true auxiliary probability~$p(z_n\given x_n, \theta)$.  This proposal can be efficiently made for each data point, but it is only necessary to evaluate~$p(z_n\given x_n, \theta)$ -- and therefore the likelihood function -- for the subset of data points which are proposed to change state.  That is, if a sample from the proposal distribution sends~${z_n = 0}$ to~${z_n = 0}$ then it doesn't matter whether we accept or reject.  If we use samples from a geometric distribution to choose the data points, it is not even necessary to explicitly sample all of the~$N$ proposals.

The probabilities $q_{b\ra d}$ and $\qq$ can be tuned as hyperparameters. If they are larger than~$p(z_n=0\given x_n, \theta)$ and~$p(z_n=1\given x_n, \theta)$ respectively, then we obtain near-perfect Gibbs sampling. But larger value also require more likelihood evaluations per iteration. Since the likelihoods of the bright data point have already been evaluated in the course of the Markov step in~$\th$ we can reuse these values and set~${q_{b\ra d} = 1}$, leaving~$\qq$ as the only hyperparameter, which we can set to something like $M/N$. The resulting algorithm, which we call \emph{implicit resampling}, is shown as Algorithm~\ref{MH_resample}.

\begin{algorithm*}[ht!]
\caption{Implicit $z_n$ sampling}
\label{MH_resample}
  \begin{algorithmic}[1]
    \For{$n \gets 1\ldots N$}
        \Comment{Loop over all the auxiliary variables.}
        \If{$z_n = 1$}
           \Comment{If currently bright, propose going dark.}
              \State $u \sim \text{Uniform}(0,1)$
            \Comment{Sample the MH threshold.}
            \If{$\displaystyle \frac{q_{d\to b}}{\tilde{L}_n(\theta)} > u$}
              \Comment{Compute MH ratio with $\tilde{L}_n(\theta)$ cached from $\theta$ update.}
               \State $z_n\gets 0$
                   \Comment{Flip from bright to dark.}
          \EndIf
    \Else
        \Comment{Already dark, consider proposing to go bright.}
        \If{$v < q_{d \to b}$ \textbf{where} $v \sim \text{Uniform}(0,1)$}
        \Comment{Flip a biased coin with probability $q_{d \to b}$.}
            \State $u \sim \text{Uniform}(0,1)$
                \Comment{Sample the MH threshold.}
            \If{$\displaystyle \frac{\tilde{L}_n(\theta)}{q_{d \to b}} < u$}
                \Comment{Compute MH ratio.}
                \State $z_n \gets 1$
                    \Comment{Flip from dark to bright.}                
            \EndIf
        \EndIf
    \EndIf
    \EndFor
  \end{algorithmic}
\end{algorithm*}

\subsection{Data structure for brightness variables}

In the algorithms shown so far, we have aimed to construct a valid Markov chain while minimizing the number of likelihood evaluations, on the (reasonable) assumption that likelihood evaluations dominate the computational cost. However, the algorithms presented do have some steps which appear to scale linearly with~$N$, even when~$M$ is constant. These are steps such as ``loop over the bright data points'' which takes time linear in~$N$.  With a well-chosen data structure for storing the variables~$z_n$, we can ensure that these operations only scale with~$M$.

The data structure needs to store the values of~$z_n$ for all~$n$ from~$1$ to~$N$, and it needs to support the following methods in~$O(1)$ time:
\begin{itemize}
  \item $\tt{Brighten}$($n$) : Set ${z_n = 1}$
  \item $\tt{ithBright}$($i$) : Return $n$, the $i$th bright data point (in some arbitrary ordering).
\end{itemize}
We similarly require $\tt{Darken}$ and $\tt{ithDark}$. The data structure should also keep track of how many bright data points there are.

\begin{figure}[t!]
\includegraphics[clip=true,trim=0 0 0 0]{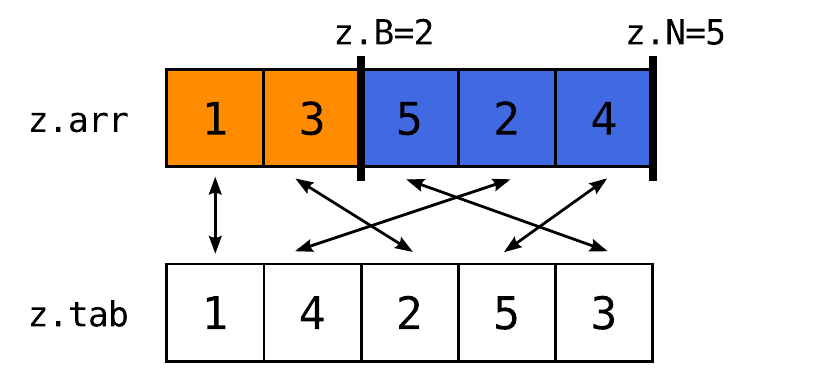}
\caption{Illustration of a data structure allowing for efficient operations on the sets of bright  and dark data points. Data points 1 and 3 are bright, the rest are dark.}
\label{z_array}
\end{figure}

To achieve this, we use the cache-like data structure shown in Figure \ref{z_array}. We store two arrays of length $N$. The first is $\tt{z.arr}$, which contains a single copy of each of the indices $n$ from $1$ to $N$. All of the bright indices appear before the dark indices. A variable $\tt{z.B}$ keeps track of how many bright indices there are, and thus where the bright-dark transition occurs. In order to also acheive $O(1)$ assignment of indices, we also maintain a direct lookup table $\tt{z.tab}$ whose $n$th entry records the position in array $\tt{z.arr}$ where $n$ is held. $\tt{Brighten}$($n$) works by looking up int $\tt{z.tab}$ the position of $n$ in $\tt{z.arr}$, swapping it with the index at position $\tt{z.B}$, incrementing $\tt{z.B}$, and updating $\tt{z.tab}$ accordingly.

\section{EXPERIMENTS}
\label{sec:experiments}

For FlyMC to be a useful algorithm it must be able to produce effectively independent samples from posterior distributions more quickly than regular MCMC. We certainly expect it to iterate more quickly than regular MCMC since it evaluates fewer likelihoods per iteration. But we might also expect it to mix more slowly, since it has extra auxiliary variables. To see whether this trade-off works out in FlyMC's favor we need to know how much faster it iterates and how much slower it mixes. The answer to the first question will depend on the data set and the model. The answer to the second will depend on these too, and also on the choice of algorithm for updating $\theta$.

We conducted three experiments, each with a different data set, model, and parameter-update algorithm, to give an impression of how well FlyMC can be expected to perform. In each experiment we compared FlyMC, with two choices of bound selection, to regular full-posterior MCMC. We looked at the average number of likelihoods queried at each iteration and the number of effective samples generated per iteration, accounting for autocorrelation. The results are summarized in Figure \ref{fig:results} and Table \ref{results_table}. The broad conclusion is that FlyMC offers a speedup of at least one order of magnitude compared with regular MCMC if the bounds are tuned according to a MAP-estimate of $\theta$. In the following subsections we describe the experiments in detail.

\subsection{Logistic regression}

We applied FlyMC to the logistic regression task described in \cite{Welling2011} using the Jaakkola-Jordan bounds described earlier. The task is to classify MNIST 7s and 9s, using the first 50 principal components (and one bias) as features. We used a Gaussian prior over the weights and chose the scale of that prior by evaluating performance on a held-out test set. To sample over~$\theta$, we used symmetric Metropolis-Hasting proposals, with step size chosen to yield an acceptance rate of 0.234 \citep{gelman-1997-accept}, optimized for each algorithm separately. We sampled the~$z_n$ using the implicit Metropolis-Hastings sampling algorithm.

We compared three different algorithms: regular MCMC, untuned FlyMC, and MAP-tuned FlyMC. For untuned FlyMC, we chose~${\xi=1.5}$ for all data points. To compute the bounds for the MAP-tuned algorithm, we performed stochastic gradient descent optimization to find a set of weights close the the MAP value and gave each data point its own~$\xi$ to make the bounds tight at the MAP parameters:~${L_n(\th_\text{MAP} )= B_n(\th_\text{MAP})}$ for all~$n$. For untuned FlyMC, and MAP-tuned FlyMC we used~${q_{d\to b}=0.1}$ and ${q_{d\to b}=0.01}$ respectively, chosen to be similar to the typical fraction of bright data points in each case.

The results are shown in Figure \ref{fig:mnist} and summarized in Table \ref{results_table}. On a per-iteration basis, the FlyMC algorithms mix and burn-in more slowly than regular MCMC by around a factor of two, as illustrated by the autocorrelation plots. Even on a per-likelihood basis, the na\"{i}ve FlyMC algorithm, with a fixed~$\xi$, performs worse than regular MCMC, by a factor of 0.7, despite needing fewer likelihood evaluations per iteration. The MAP-tuned algorithm was much more impressive: after burn-in, it queried only 207 of the 12,2214 likelihoods per iteration on average, giving a speedup of more than 20, even taking into account the slower per-iteration mixing time. We initialized all chains with draws from the prior. Notice that the MAP-tuned algorithm performs poorly during burn-in, since the bounds are less tight during this time, whereas the reverse is true for the untuned algorithm.

\begin{table*}[t!]
    \resizebox{\textwidth}{!}{%
  \begin{tabular}{ rl c c c c}
      \toprule
     &  &            & Average            &  Effective           & Speedup     \\
     &  & Algorithm  & Likelihood queries &  Samples per         & relative to  \\
     &  &            & per iteration      &  1000 iterations     & regular MCMC \\
     \midrule
    Data set: & MNIST                 & Regular MCMC    & 12,214     & 3.7    & (1) \\
    \cmidrule(r){3-6}
       Model: & Logistic regression   & Untuned FlyMC   &  6,252     & 1.3    & 0.7 \\
    \cmidrule(r){3-6}
     Updates: & Metropolis-Hastings   & MAP-tuned FlyMC &    207     & 1.4    & 22  \\
    \midrule
    Data set: & 3-Class CIFAR-10      & Regular MCMC    & 18,000     & 8.0    & (1)  \\
    \cmidrule(r){3-6}
       Model: & Softmax classifcation & Untuned FlyMC   &  8,058     & 4.2    & 1.2  \\
    \cmidrule(r){3-6}
     Updates: & Langevin  & MAP-tuned FlyMC             &    654     & 3.3    & 11   \\
    \midrule
    Data set: & OPV                   & Regular MCMC    & 18,182,764 & 1.3 & (1) \\
    \cmidrule(r){3-6}
       Model: & Robust regression     & Untuned FlyMC   &  2,753,428 & 1.1 & 5.7   \\
    \cmidrule(r){3-6}
     Updates: & Slice sampling        & MAP-tuned FlyMC &    575,528 & 1.2 & 29  \\
    \bottomrule

  \end{tabular}
  }
  \caption{Results from empirical evaluations.  Three experiments are shown: logistic regression applied to MNIST digit classification, softmax classification for three categories of CIFAR-10, and robust regression for properties of organic photovoltaic molecules, sampled with random-walk Metropolis--Hastings, Langevin-adjusted Metropolis, and slice sampling, respectively.  For each of these, the vanilla MCMC operator was compared with both untuned FlyMC and FlyMC where the bound was determined from a MAP estimate of the posterior parameters.  We use likelihood evaluations as an implementation-independent measure of computational cost and report the number of such evaluations per iteration, as well as the resulting sample efficiency (computed via R-CODA \citep{coda}), and relative speedup.}
\label{results_table}
\end{table*}

\subsection{Softmax classification}

Logistic regression can be generalized to multi-class classification problems by softmax classification. The softmax likelihood of a data point belonging to class $k$ of $K$ classes is

\begin{equation*}
L_n(\theta) = \frac{\exp(\th_k^{\trans} x_n)}{\sum_{k'=1}^K\exp(\th_{k'}^{\trans} x_n)}
\end{equation*}

Where $\th$ is now a ${K \times D}$ matrix. The Jaakkola-Jordan bound does not apply to this softmax likelihood, but we can use a related bound, due to \citet{bohning1992multinomial}, whose log matches the value and gradient of the log of the softmax likelihood at some particular~$\th$, but has a tighter curvature. \citet{MurphyBook} has the result in full in the chapter on variational inference.

We applied softmax classification to a three-class version of CIFAR-10 (airplane, automobile and bird) using 256 binary features discovered by \citet{krizhevsky-2009a} using a deep autoencoder. Once again, we used a Gaussian prior on the weights, chosen to maximize out-of-sample performance. This time we used the Metropolis-adjusted Langevin algorithm (MALA, \citet{roberts1996exponential}) for our parameter updates. We chose the step sizes to yield acceptance rates close to the optimal 0.57 \citep{roberts1998optimal}.  Other parameters were tuned as in the logistic regression experiment.

The softmax experiment gave qualitatively similar results to the logistic regression experiment, as seen in Figure~\ref{fig:cifar} and Table~\ref{results_table}. Again, the MAP-tuned FlyMC algorithm dramatically outperformed both the lackluster untuned FlyMC and regular MCMC, offering an 11-fold speedup over the latter.

\subsection{Robust sparse linear regression}

Linear regression with Gaussian likelihoods yields a closed-form expression for the posterior. Non-Gaussian likelihoods, however, like heavy-tailed distributions used in so-called ``robust regression'' do not. Our final experiment was to perform inference over robust regression weights for a very large dataset of molecular features and computed electronic properties. The data set, described by \citet{cep} consists of 1.8 million molecules, with 57 cheminformatic features each, and the task was to predict the HOMO-LUMO energy gap, which is useful for predicting photovoltaic efficiency.

We used a student-t distribution with~${\nu=4}$ for the likelihood function and we computed a Gaussian lower bound to this by matching the value and gradient of the t distribution probability density function value at some $\xi$ (${\xi=0}$ for the untuned case,~${\xi = \th_{MAP}^{\trans}}x$ for the MAP-tuned case). We used a sparsity-inducing Laplace prior on the weights. As before, we chose the scale of the prior, and of the likelihood too, to optimize out-of sample performance.

\begin{figure*}[t]
    \centering%
    \subfloat[MNIST with MH]{%
        \includegraphics[width=0.32\textwidth]{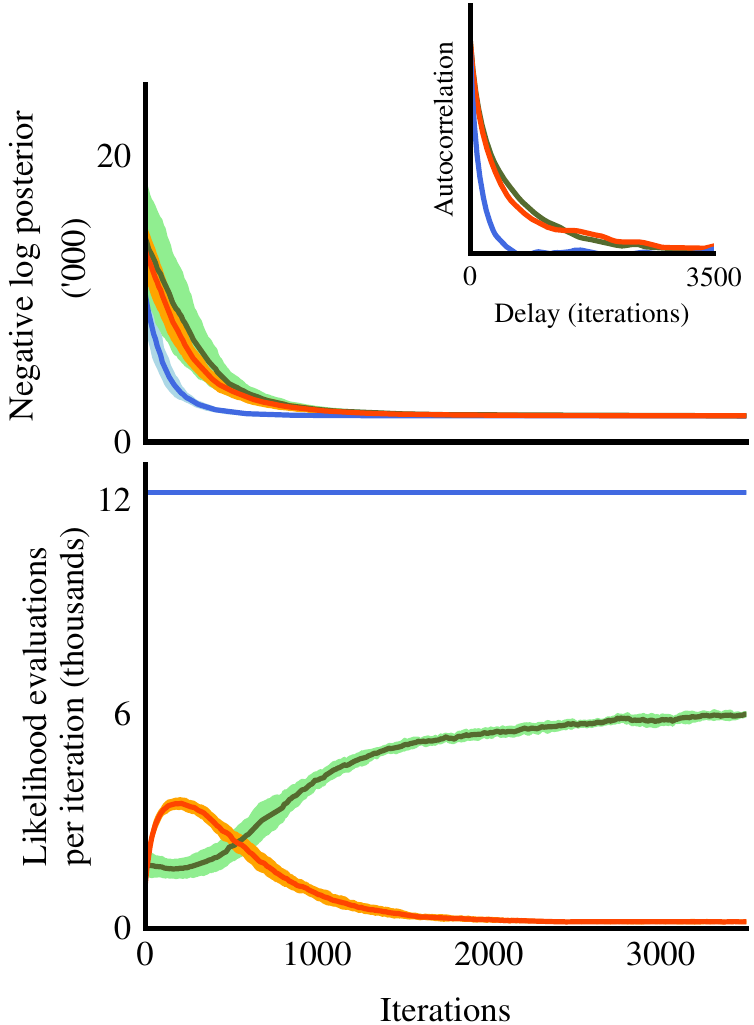}
        \label{fig:mnist}
        }~
    \subfloat[CIFAR-10 with Langevin]{%
        \includegraphics[width=0.32\textwidth]{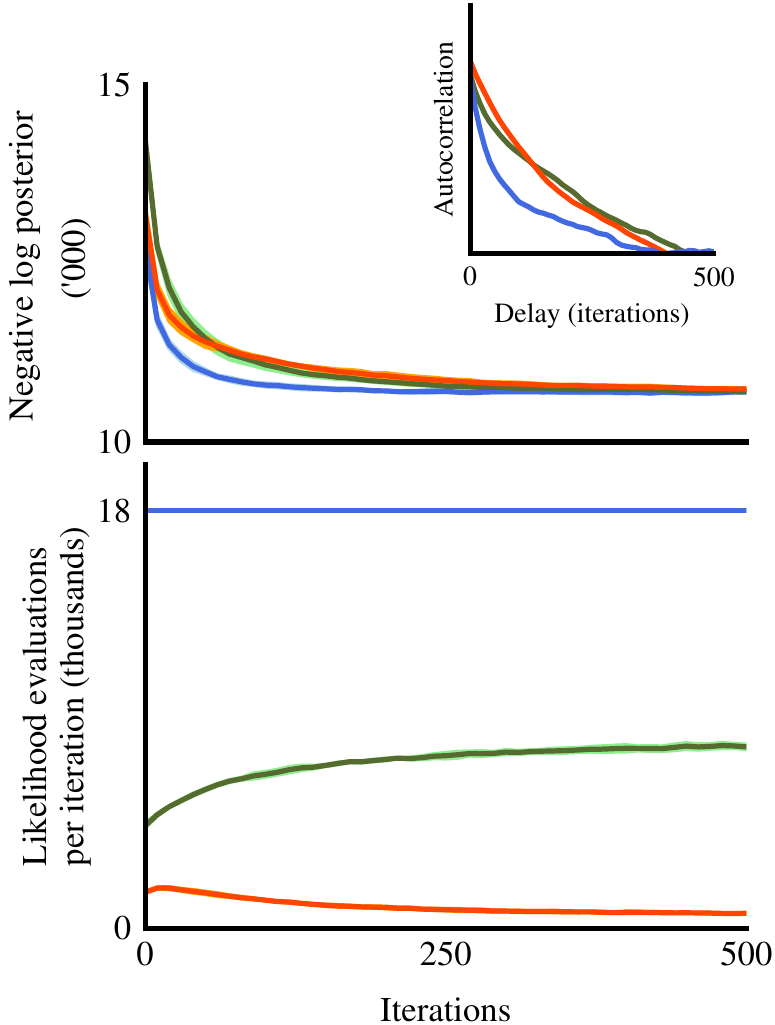}
        \label{fig:cifar}
        }~
    \subfloat[OPV with Slice Sampling]{%
        \includegraphics[width=0.32\textwidth]{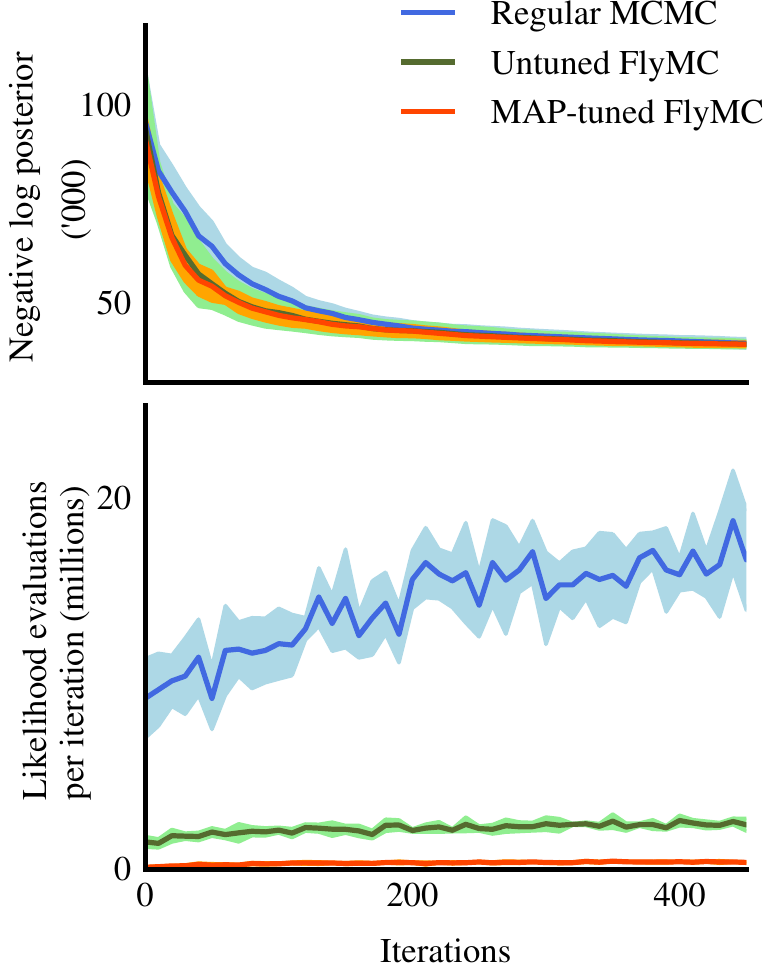}
        \label{fig:opv}
        }
    \caption{Tuned and untuned Firefly Monte Carlo compared to regular MCMC with three different operators, data sets, and models: (a)~the digits 7 and 9 from the MNIST data are classified using logistic regression, with a random-walk Metropolis-Hastings operator; (b)~softmax classification on three classes (airplane, automobile, and bird) from the CIFAR-10 image dataset, using Langevin-adjusted Metropolis; (c)~robust regression on the HOMO-LUMO gap (as computed by density functional theory calculations) for a large set of organic photovoltaic molecules, using slice sampling.  In each subfigure, the top shows the trace of the log posterior density to illustrate convergence, and the bottom shows the average number of likelihoods computed per iteration.  One standard deviation is shown around the mean value, as computed from five runs of each.  The blue lines are computed using the full-data posterior, and the green and orange lines show the untuned and tuned Firefly MC traces, respectively.}
    \label{fig:results}
\end{figure*}

We performed parameter updates using slice sampling \citep{Neal03}. Note that slice sampling results in a variable number of likelihood evaluations per iteration, even for the regular MCMC algorithm. Again, we found that MAP-tuned FlyMC substantially outperformed regular MCMC, as shown in Figure~\ref{fig:opv} and Table~\ref{results_table}.

\section{DISCUSSION}
\label{sec:discussion}

In this paper, we have presented Firefly Monte Carlo, an algorithm for performing Markov chain Monte Carlo using subsets (minibatches) of data.  Unlike other recent proposals for such MCMC operators, FlyMC is exact in the sense that it has the true full-data posterior as its target distribution.  This is achieved by introducing binary latent variables whose states represent whether a given datum is bright (used to compute the posterior) or dark (not used in posterior updates).  By carefully choosing the conditional distributions of these latent variables, the true posterior is left intact under marginalization.  The primary requirement for this to be efficient is that the likelihoods term must have lower bounds that collapse in an efficient way.

There are several points that warrant additional discussion and future work.  First, we recognize that useful lower bounds can be difficult to obtain for many problems.  It would be useful to produce such bounds automatically for a wider class of problems.  As variational inference procedures are most often framed in terms of lower bounds on the marginal likelihood, we expect that Firefly Monte Carlo will benefit from developments in so-called ``black box'' variational methods \citep{wang2013variational, Ranganath2014}.  Second, we believe we have only scratched the surface of what is possible with efficient data structures and latent-variable update schemes.  For example, the MH proposals we consider here for~$z_n$ have a fixed global~$q_{d\to b}$, but clearly such a proposal should vary for each datum.  Third, it is often the case that larger state spaces lead to slower MCMC mixing.  In Firefly Monte Carlo, much like other auxiliary variable methods, we have expanded the state space significantly.  We have shown empirically that the slower mixing is more than offset by the faster per-transition computational time.  In future work we hope to show that fast mixing Markov chains on the parameter space will continue to mix fast in the Firefly auxiliary variable representation.

Firefly Monte Carlo is closely related to recent ideas in using pseudo-marginal MCMC \citep{andrieu2009pseudo} for sampling from challenging target distributions. If we sampled each of the variables $\{z_n\}$ as a Bernoulli random variable with success probability 0.5, then the joint posterior we have been using becomes un unbiased estimator of the original posterior over $\th$, up to normalization. Running pseudo-marginal MCMC using this unbiased estimator would be a special case of FlyMC: namely FlyMC with $z$ and $\th$ updated simultaneously with Metropolis-Hastings updates.

\subsubsection*{Acknowledgements}
Thanks to Andrew Miller, Jasper Snoek, Michael Gelbart, Brenton Partridge, and Elaine Angelino for helpful discussions.  Partial funding was provided by Analog Devices (Lyric Labs) and DARPA Young Faculty Award N66001-12-1-4219.

\bibliography{refs}
\bibliographystyle{plainnat}

\end{document}